\documentclass{article}

\usepackage[round]{natbib}
\usepackage[utf8]{inputenc} 
\usepackage[T1]{fontenc}    
\usepackage{hyperref}       
\usepackage{url}            
\usepackage{booktabs}       
\usepackage{amsfonts}       
\usepackage{nicefrac}       
\usepackage{microtype}      
\usepackage{amsmath,amssymb,amsthm,amsfonts}
\usepackage{times}
\usepackage{xr}
\usepackage{hyperref}
\usepackage{url}
\usepackage{amssymb}
\usepackage{amsbsy}
\usepackage{amsthm}
\usepackage{amsmath}
\usepackage{paralist}
\usepackage{bm}
\usepackage{booktabs}
\usepackage{rotating}
\usepackage{cases}
\usepackage{mathtools}
\usepackage{graphicx}
\usepackage{algorithm}
\usepackage{algpseudocode}
\usepackage{setspace}
\usepackage{textcomp}

\newcommand{\vnorm}[1]{\left|\left|#1\right|\right|}
\newcommand{\R}{\mathbb{R}}
\title{Better Conditional Density Estimation for Neural Networks}

\author{
  Wesley Tansey \\
  Department of Computer Science\\
  University of Texas at Austin\\
  Austin, TX 78722 \\
  \texttt{tansey@cs.utexas.edu} \\
  \and
  Karl Pichotta \\
  Department of Computer Science\\
  University of Texas at Austin\\
  Austin, TX 78722 \\
  \texttt{kpich@cs.utexas.edu} \\
  \and
  James G.~Scott \\
  Department of Statistics and Data Science\\
  University of Texas at Austin\\
  Austin, TX 78722 \\
  \texttt{james.scott@mccombs.utexas.edu} \\
}

\begin{document}

\maketitle

\begin{abstract}
The vast majority of the neural network literature focuses on predicting point values for a given set of response variables, conditioned on a feature vector. In many cases we need to model the full joint conditional distribution over the response variables rather than simply making point predictions. In this paper, we present two novel approaches to such conditional density estimation (CDE): Multiscale Nets (MSNs) and CDE Trend Filtering. Multiscale nets transform the CDE regression task into a hierarchical classification task by decomposing the density into a series of half-spaces and learning boolean probabilities of each split. CDE Trend Filtering applies a $\text{k}^{\text{th}}$ order graph trend filtering penalty to the unnormalized logits of a multinomial classifier network, with each edge in the graph corresponding to a neighboring point on a discretized version of the density. We compare both methods against plain multinomial classifier networks and mixture density networks (MDNs) on a simulated dataset and three real-world datasets. The results suggest the two methods are complementary: MSNs work well in a high-data-per-feature regime and CDE-TF is well suited for few-samples-per-feature scenarios where overfitting is a primary concern.
\end{abstract}

\section{Introduction}
\label{sec:introduction}

In the last decade, deep neural networks have been at the core of many state-of-the-art machine learning systems due to their exceptional ability to learn complicated, non-linear functions of large dimension. When employed to solve real- and ordinal-valued regression problems, almost invariably such networks are trained to produce a point estimate.  But often an interval estimate (i.e.~a prediction interval) is necessary.  One na\"ive approach is to simply base a predictive error bar using the root mean-squared error of the network.  But this is rarely sensible in practice: the \textit{conditional} predictive uncertainty of the network is likely to depend strongly on the features used to train the model.  In the statistics literature, this is referred to as \textit{conditional heteroskedasticity}: the variance of the model residuals is itself a function of the features.  There is a pressing need for methods which produce sensible interval predictions from deep nets.

If a user wishes instead to infer a conditional density rather than a point, their options are typically one of the following.
\begin{enumerate}
\item Discretize the variable and model it using a multinomial classifier. While this is fast, it destroys the underlying topological structure of the variable's underlying space by making each bin independent.  It therefore leads to ``lumpy'' density estimates and reduces sample efficiency due to high variance.

\item Make a parametric assumption about the form of the conditional density, such as a fixed-size Gaussian mixture model \citep[also called Mixture Density Networks, see][]{bishop:mdn:1994} or Gaussian-Pareto mixtures \citep{carreau2009hybrid}, and build a model for conditional parameters of that parametric distribution. When increasing in dimensionality of the target variable, this may require making independence assumptions in order to keep the covariance estimations tractable.

\item Add dropout at inference time \citep{gal2015dropout}. This works well for measuring uncertainty about one's point estimate.  But sampling uncertainty about a maximum likelihood point estimate is not the same as modeling the distribution of outcomes; the latter is typically much wider.

\item Use a Bayesian deep learning framework. While much work in this area is just emerging \citep[e.g.][]{pu2015deep}, many of the existing architectures, such as LSTMs, do not yet have a Bayesian interpretation. Furthermore, posterior inference on such models can be prohibitively expensive in the case where billions of evaluations must be performed, as in reinforcement learning contexts, for example.
\end{enumerate}
Thus all four of the above options are lacking in some crucial way that prevents them from being used in practice.

In this paper we seek to overcome these issues by presenting two approaches to conditional density estimation that are nonparametric, scalable, make no independence assumptions, and leverage the underlying topological structure of the variables.  Our first approach, Multiscale Nets, decomposes the density into a series of half-spaces via a dyadic decomposition.  This is the more flexible of our two models; it essentially turns density estimation into hierarchical classification, and it is designed to be a maximally flexible model for situations with a favorable ratio of the number of samples to the feature-set size.  Our second approach, CDE trend filtering (TF), couples a multinomial model (like option 1 above) with a trend-filtering penalty to introduce smoothness in the underlying density estimate.  Because this incorporates additional regularization compared to the multiscale case, we envision it as a better approach in situations where data is sparser as a function of the number of features.  In each case, the features are mapped to raw logits---binomial in the multiscale case, multinomial in the CDE TF case---via an appropriate neural network. This paper presents extensive evidence that each of these two methods is superior to Gaussian mixture models (the current state of the art) in its appropriate domain.

\section{Multiscale nets}
\label{sec:multiscale}

\begin{figure}
\includegraphics[width=0.9\textwidth]{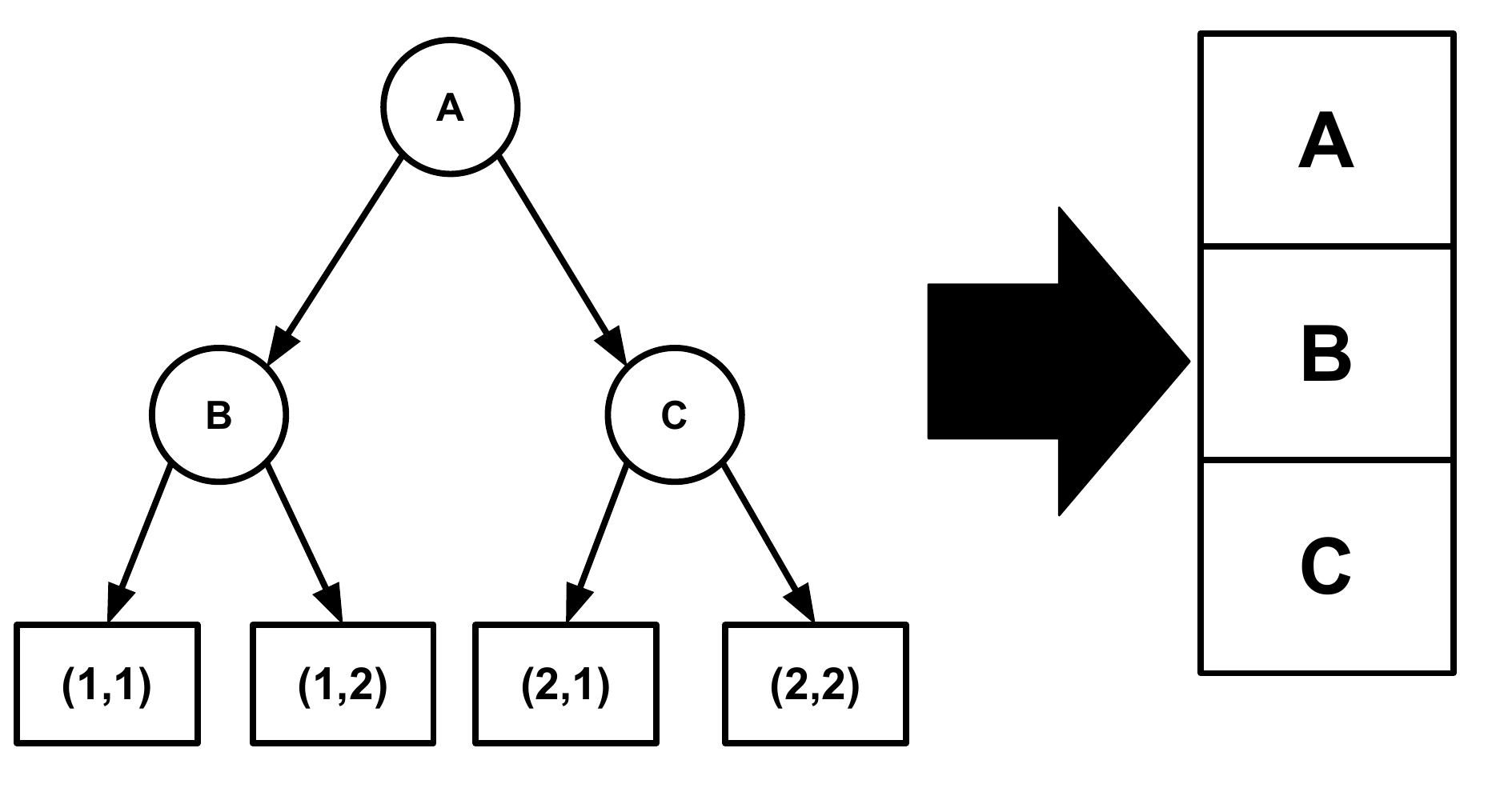}
\caption{\label{fig:multiscale-decomposition} The multiscale decomposition visualized. Every dimension in the response variable is iteratively divided into half spaces and every split becomes an output node in the network. A given example then has $\log_2(p)$ labels for discrete density with $p$ bins.}
\end{figure}

\subsection{Dyadic partitions}

We use $F$ to denote a probability measure on $B$, $f$ the corresponding density function, and $F(A) = \int_A dF$ the probability of set $A \subset B$.  Our approach to conditional density estimation relies upon constructing a recursive dyadic partition of $B$.  The level-$k$ partition, denoted $\Pi^{(k)}$, via a bijection between $\Pi^{(k)}$ and all length-$k$ binary sequences $\gamma \in \{0,1\}^k$, as follows.  Let the level-$1$ partition as $\Pi^{(1)} = \{B_0, B_1\}$ where $B_0 \cup B_1 = B$ and $B_0 \cap B_1 = \emptyset$.  Given the partition at level $k$, the level $k+1$ partition is constructed by specifying, for all $\gamma \in \{0,1\}^k$, a pair $(B_{\gamma 0}, B_{\gamma 1})$ such that $B_{\gamma 0} \cup B_{\gamma 1} = B_{\gamma}$ and $B_{\gamma 0} \cap B_{\gamma 1} = \emptyset$.  Here $\gamma0$ (or $\gamma1$) is new binary sequence defined by appending a 0 (or 1) to the end of $\gamma$. If $\gamma$ is an empty string, then $B_{\gamma}$ is the root node, i.e.~$B$.  For example, if $B$ is the unit interval (i.e.~the level-0 partition), the level-$1$ partition could be $\{[0,0.5], (0.5, 1]\}$; the level-$2$ partition could be
$$
\Pi^{(2)} = \{[0, 0.25], (0.25, 0.5], (0.5, 0.75], (0.75, 1]\} \, ;
$$
and so on.  We refer to $B_{\gamma}$ as a parent node, to $B_{\gamma0}$ as the left child, and to $B_{\gamma1}$ as the right child.

Suppose that $Y \sim F$ is a draw from $F$.  We characterize the probability measure $F$ via the conditional ``splitting'' probabilities
$$
w_{\gamma} = P(Y \in B_{\gamma0} \mid Y \in B_{\gamma}) \, ;
$$
that is, the probability that the $Y$ will fall in the left-child set, given that it falls in the parent set.  Because $B_{\gamma0} \subset B_{\gamma}$ and therefore $Y \in B_{\gamma0} \implies Y \in B_{\gamma}$, we have the following representation for $w_{\gamma}$:
\begin{eqnarray}
P(Y \in B_{\gamma0}) &=& P(Y \in B_{\gamma0}, Y \in B_{\gamma}) \nonumber \\
 &=& P(Y \in B_{\gamma0} \mid Y \in B_{\gamma}) \cdot P(Y \in B_{\gamma}) \nonumber \\
 &=& w_{\gamma} \cdot P(Y \in B_{\gamma}) \label{eqn:tree_recursion_probability} \, .
\end{eqnarray}
Thus $w_{\gamma}$ is given by the ratio of probabilities
$$
w_{\gamma} \equiv P(Y \in B_{\gamma0})  / P(Y \in B_{\gamma}) = F(B_{\gamma0}) / F(B_{\gamma}) \, .
$$
Moreover, suppose we apply Equation (\ref{eqn:tree_recursion_probability}) recursively to itself, i.e.~to $P(Y \in B_{\gamma})$ on the right-hand side, and proceed up the tree until arriving at the root node $B$ (for which $P(Y \in B) = 1$).  This allows us to express the probabilities at the terminal nodes of the tree, which form a discrete approximation to the probability density function, as the product of splitting probabilities $w_{\gamma}$ as one traverses up the tree to the root node.

\subsection{Incorporating features}

We incorporate features as follows.  Let $\mathcal{X}$ denote a feature space, and let $F_{x}$ for $x \in \mathcal{X}$ denote a probability measure over $B$ specific to $x$.  (We assume that all $F_x$ have the same support.)  Our approach to multiscale conditional density estimation is to allow the conditional splitting probabilities in the dyadic partition to depend upon $x$ via the logistic transform of some function $r_\gamma$.  Specifically, we let
$$
w_{\gamma}(x) = P(Y \in B_{\gamma0} \mid Y \in B_{\gamma}, x) = \frac{ \exp\{r_\gamma(x)\}} {1 + \exp\{r_\gamma(x) \} } \, .
$$
This turns the problem of density estimation into a set of independent classification problems: for every $\gamma$, we learn a function $r_{\gamma}(x)$ that predicts how likely that an outcome $y$ that falling in the parent node $B_{\gamma}$ will also fall in the left-child node $B_{\gamma 0}$.

\subsection{Related work}

There is a significant body of work in statistics on conditional density estimation.  Most frequentist work on this subject is based on kernel methods \citep[see, e.g.][and the references contained therein]{bash:hynd:2001}.  But traditional kernel methods do poorly at estimating densities which contain both spiky and smooth features ,and which require adaptivity to large jumps both.  Moreover, conditional density estimation using kernel methods requres the estimation of a potentially high-dimensional joint density $p(y,x)$ as a precursor to estimation $p(y \mid x)$.  We avoid the difficult task of estimating $p(y,x)$, focusing on $p(y \mid x)$ directly. 

Multiscale nets essentially aim to treat conditional density estimation as a hierarchical classification problem. A similar approach for doing one-dimensional CDE has been proposed by \citep{stone:etal:2003} via boosting machines in a manner similar to ordinal regression. However, their approach requires heuristics to deal with a monotonicity requirement in their decomposition bins. In the neural network literature, a very similar technique has been used in neural language models \citep{morin2005hierarchical}. Their results build on ours and our dyadic decomposition could equally be data adaptive, in the case where the depth of the tree is limited, by simply choosing splits via percentiles of the distribution.

This device for exploiting the conditional-independence properties of a tree is also used to define a P\'olya-tree prior and other kinds of multiscale methods in nonparametric Bayesian inference \citep{mauldin:sudderth:williams:1992,ma:2014}.  Here, a random probability measure $F$ is constructed by assuming that the conditional probability $w_{\gamma} = F(B_{\gamma0}) / F(B_{\gamma})$ is a different beta random variable for each node $\gamma$ in an infinitely deep tree.  The parameters of each beta random variable are determined by a concentration parameter $\alpha$ and a base measure $F_0$.  It is also similar to multiscale models for Poisson intensity estimation \citep{fryzlewicz:nason:2004,jansen:2006,willett:nowak:2007}.   Our approach differs in that we incorporate covariates into the spitting probabilities, and in that we do not work explicitly within the Bayesian formalism (i.e.~by placing a prior over the space of probability measures).

\section{CDE Trend Filtering}

In this section we define a ``flat'' (i.e.~non-hierarchical) version of a conditional density estimator via neural nets.  To do so, we generalize recent advances in trend filtering, a nonparametric method for regression and smoothing over graphs.  Graph trend filtering \citep{wang2014trend} minimizes the following objective:
\begin{equation}
\label{eqn:graph-trend-filtering}
\underset{\beta \in \R^n}{\text{minimize}} \quad l(\beta) + \lambda \vnorm{\Delta \beta}_1 \, ,
\end{equation}
where $l$ is a smooth, convex loss function. Here $\Delta$ is the $k^{\text{th}}$-order trend filtering penalty matrix, where the $k=0$ base matrix is the oriented edge matrix encoding the relationship between the elements of $\beta$. The resulting $\ell_1$ regularization term aims to drive the $k^{\text{th}}$-order differences between the $\beta$'s to zero.

We define conditional density estimation trend filtering (CDE-TF) as follows.  Let $\Pi = (I_1, \ldots, I_D)$ be a set of (possibly multivariate) histogram bins, i.e.~a flat partition of $B$, the support of the underlying probability measure.  We use $c_i$ as a bin indicator for the response variable $y_i$: that is, $c_i = j$ if $y_i \in I_j$.  In CDE-TF, we model the $c_i$'s directly as categorical random variables, where the the probabilities $P(c_i = j \mid x_i)$ depend on features via the softmax function:
\begin{equation}
\label{eqn:categorical_model}
(c_i \mid x_i) \sim \mbox{Categorical}(\eta(x)) \, , \quad \mbox{where} \quad \eta_j(x) =
\frac{\exp[\psi_j(x)]}
{\sum_{l=1}^D \exp[\psi_l(x)]} \, .
\end{equation}
To parametrize and regularize the $\psi_j(x)$'s, we combine two approaches:
\begin{enumerate}
\item We set the $\psi_j(x)$'s to be the output of an appropriate neural network.
\item We apply a graph trend-filtering penalty directly to these outputs, by penalizing the quantity $\Vert \Delta \psi(x) \Vert_1$ where $\psi(x)$ is the stacked vector of outputs $\psi_j(x)$ from the network and $\Delta$ is the trend-filtering penalty matrix.  Here the graph used to construct $\Delta$ is determined by the adjacency structure of the bins $I_1, \ldots, I_D$.  In the vast majority of all cases, this graph will simply be a $K$-dimensional grid graph, where $K$ is the dimension of the response vector $y$.
\end{enumerate}

Thus the objective we are minimizing is
\begin{equation}
\label{eqn:cde-trend-filtering}
{\text{minimize}} \quad \sum_{i} l_i(\psi(x_i)) + \lambda \vnorm{\Delta \psi(x) }_1 \, ,
\end{equation}
where $l_i(\psi(x_i))$ is the contribution to the loss function from Model (\ref{eqn:categorical_model}) for the $i^{\text{th}}$ response, $\psi(x)$ is the network output that returns returns raw logits, and $x_i$ is the $i^{\text{th}}$ training example. Throughout this paper, we assume that $\psi$ is a neural network, but any differentiable function is acceptable (so that the domain of minimization will be context-dependent). Conceptually, goal of CDE-trend filtering is to bring the representational power of neural networks to the task of density estimation, while simultaneously regularizing the model output to ensure smooth estimated densities, borrowing information across adjacent bins and effecting a favorable bias--variance trade-off.

\begin{figure}
\begin{center}
\includegraphics[width=0.9\textwidth]{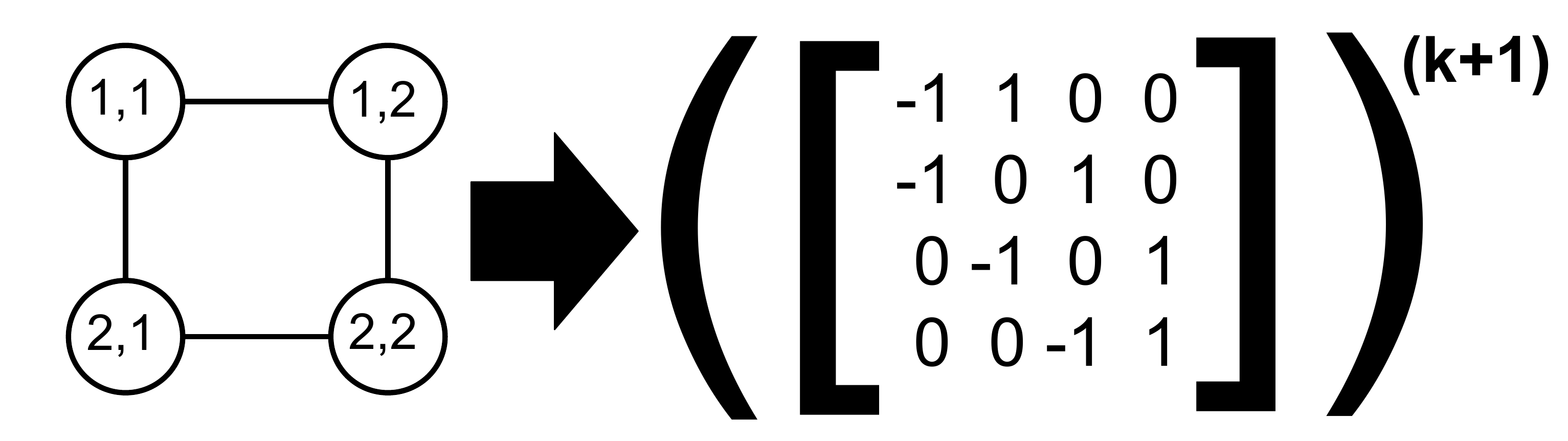}
\caption{\label{fig:trend-filtering} The graph trend filtering penalty visualized. On the left, each bin in the discrete multidimensional density has an edge to
its closest neighbor, forming a lattice, and encoded as the oriented edge matrix on the right. The $k^{\text{th}}$-order trend filtering penalty is then the matrix resulting from multiplying the matrix by itself (or its transpose if on an odd step) $k$ times.}
\end{center}
\end{figure}
\

\section{Experiments}
\label{sec:experiments}

\subsection{Setup}
\label{sec:setup}
We evaluate our two approaches against the most common CDE architectures for neural networks: plain multinomial classifiers and mixture density networks (MDNs) \citep{bishop:mdn:1994}. The latter approach corresponds to having a neural network output the parameters of a Gaussian mixture model (GMM). Despite being more than two decades old, MDNs are still often the best-performing conditional density estimator \citep{sugiyama2010conditional} and have had recent success with deep architectures \citep{zen2014deep}. Most work on MDNs assumes either independence of the variables \citep[as in, e.g.][]{zen2014deep}, or only deals only with univariate densities. Modeling the joint density over variables with MDNs is in general much more difficult, as it requires outputting a positive semi-definite covariate matrix. We implement such a model by having it output the lower triangular entries in the Cholesky decomposition of the covariance matrix for each mixture component \citep[see][for more details on this approach]{lopes2011cholesky}. In part due to the difficulty in constructing multi-dimensional MDNs, even recent work that requires deep, multidimensional conditional density estimation \citep[e.g.][]{zhang2016colorful} resorts to using simple multinomial grids. We therefore compare against both methods as reasonable baselines; we also provide a point estimate model for RMSE comparisons.

In all of our experiments, we focus on predicting discrete densities. As such, all target variables are discretized on an evenly-spaced grid spanning their empirical range. For the one-dimensional targets, we use a grid of $32$ bins for the synthetic experiment and $128$ bins for the S-class dataset; for the two-dimensional targets, we use a $32\times32$ lattice. Performance is measured in terms of both log-probability of the test set and root mean squared error (RMSE) of each model's point estimate. For all experiments, we select the best trend filtering model based on a grid search for the best $(\lambda, k)$ pair based on log-probability on the validation set; we conduct a similar search for the number of GMM mixture components in the real-world datasets. All real-world datasets are randomly split into 80\%/10\%/10\% train/validation/test samples, with results averaged over five, five, and ten independent trials for the Parkinson's rental rates, Mercedes datasets, respectively.

\subsection{Synthetic experiment: MNIST Distributions}
\label{sec:mnist}

\begin{figure}
\includegraphics[width=0.9\textwidth]{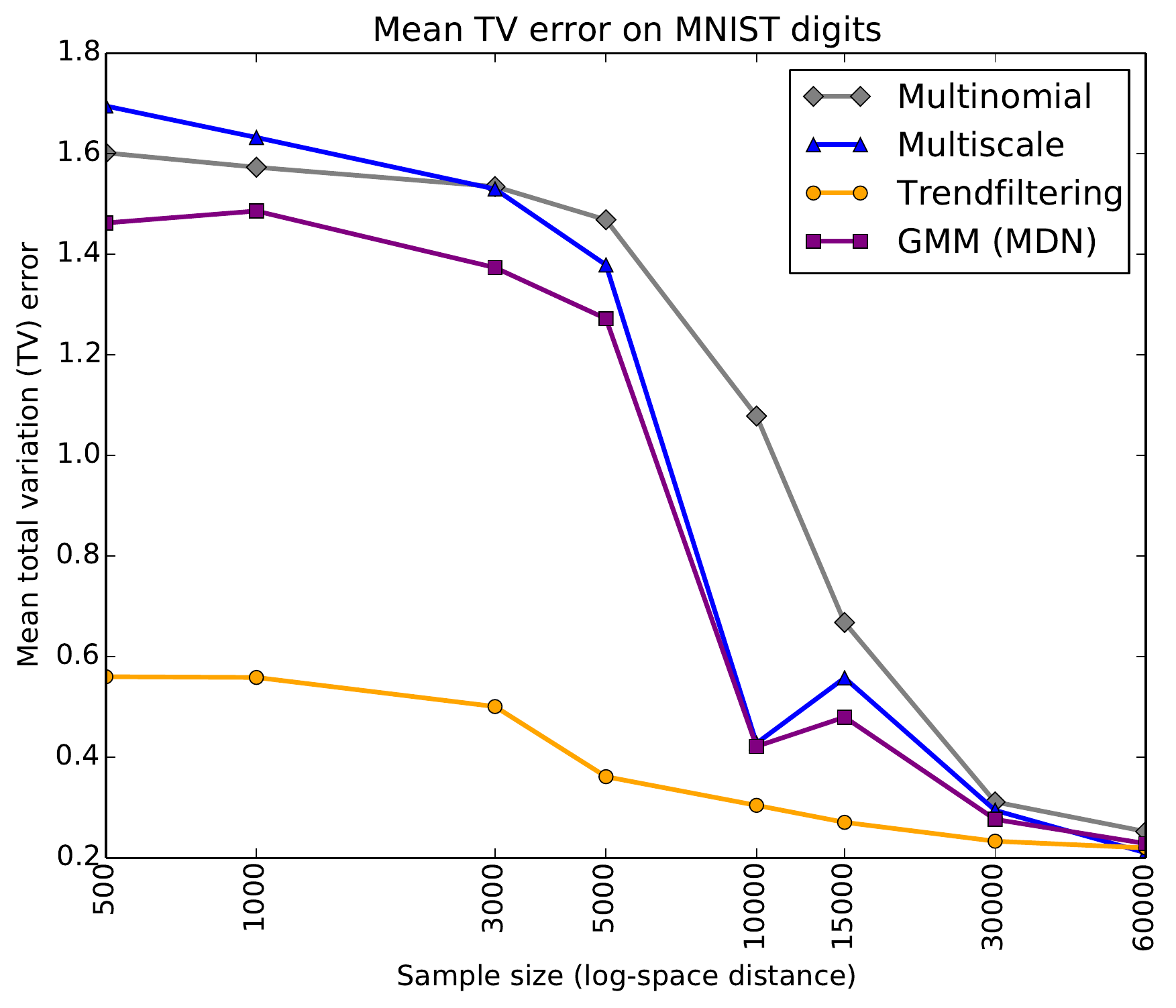}
\caption{\label{fig:mnist_tv} Performance of each CDE model on the synthetic MNIST-Distributions dataset, as a function of the number of training samples. The trend filtering method strongly outperforms the other models in the low-sample regimes.}
\end{figure}

The MNIST dataset is a well-known benchmark classification task of mapping a $28\times28$ gray scale handwritten digit to its corresponding digit class. We modify this dataset by mapping each digit class to a randomly-generated, discretized, three-component Gaussian mixture model. The digit labels are then replaced with a random draw from this density. From an investigatory perspective, this dataset is ideal for demonstrating sample efficiency, since convolutional neural networks are known to perform with over $99\%$ accuracy in the classification setting. Figure \ref{fig:mnist_tv} shows how the performance of each model improves with the number of samples. Note that at 500 samples, the model is seeing just over 1 example per bin per class, on average. In these scenarios, making some sort of assumption about the underlying conditional distribution is necessary. The adaptive piecewise-polynomial assumption made by the trend filtering method is clearly effective at fitting such multi-modal mixtures when the sample size is small. Interestingly, the trend filtering method also strongly outperforms the GMM model, despite the fact that it is parameterized with the same number of components as the underlying ground truth GMM. One possible reason for this, as we see in the real-world experiments, is the difficulty of finding a good fit for a GMM without overfitting in the small-sample case. 

\begin{table}
\begin{center}
\begin{tabular}{|l|rr|rr|rr|}
\hline
& \multicolumn{2}{|c|}{Parkinson's Scores} & \multicolumn{2}{|c|}{Mercedes Prices} & \multicolumn{2}{|c|}{Rental Rates} \\
Model              & Log(Prob) & RMSE & Log(Prob) & RMSE & Log(Prob) & RMSE \\\hline
Point estimate     & N/A & 9.92   &   N/A & 3.60 &   N/A & 4.76    \\
Multinomial        & -7.00 & 9.93 & -2.35 & 3.87 & -3.87 & 4.71    \\
GMM (MDN)          & -6.50 & 9.68 & -2.30 & 3.64 & -4.63 & 4.84    \\
Multiscale         & -6.62 & \textbf{9.41} & \textbf{-2.21} & \textbf{3.46} & \textbf{-3.82} & \textbf{4.61}    \\
Trend Filtering    & \textbf{-6.16} & 9.48 & -2.34 & 3.90 & -3.90 & 4.79     \\\hline
\end{tabular}
\caption{\label{tab:results} Performance of each CDE model on the three real-world datasets. For each dataset, we provide both the log probability of the hold-out observation and the RMSE of the point estimate. The latter shows that, surprisingly, estimating the full conditional density is either free or even benefitial compared to a point estimate model.}
\end{center}
\end{table}

\subsection{Parkinson's Disease Telemonitoring}
\label{sec:parkinsons}

The Parkinson's Telemonitoring dataset \citep{dataset:parkinsons} consists of biomedical voice measurements from 42 people with early-stage Parkinson's disease. The goal is to predict the motor and total Unified Parkinson's Disease Rating Scale (UPDRS) scores, which are highly correlated for each patient, but can exhibit stark discontinuities and multimodalities. Each patient appears in the dataset approximately 200 times, with each appearance corresponding to an example in the dataset, with an indicator variable specifying which patient is speaking, as well as 18 other real-valued features. After discretizing the two scores, the resulting problem is thus very similar to the low-sample scenario from \ref{sec:mnist}. The first column in Table \ref{tab:results} confirms that the situation is similar, with the trend filtering model performing much stronger than the other methods.

We also note that both the multiscale and trend filtering methods have RMSE scores that strongly outperform a baseline point estimate model. Thus, even in the case of modeling point predictions, it is actually beneficial for one to model the entire joint density. This result is both surprising and promising, as we are effectively seeing a free-lunch: improved point prediction that comes with predicted error bars.

\subsection{Mercedes S-class Sale Prices}
\label{sec:sclass}

The Mercedes dataset consists of sale prices for approximately 30K used Mercedes S-class sedans and fourteen features relating to the car. In contrast to the two previous experiments, this dataset is in a much higher sample-to-feature regime. Column 2 of Table \ref{tab:results} indicates that under this regime, the trend filtering smoothing is unnecessary, as it performs nearly the identically to the multinomial model. Instead, the multiscale method is now the clear best choice model, outperforming all other methods in both categories.

\subsection{Real Estate Rentals}
\label{sec:realestate}

The final benchmark dataset covers approximately 8K real estate rentals, with 14 features per property. The goal is to estimate the joint conditional density of rental price and occupancy rate. The results of the CDE models are in the third column of Table \ref{tab:results}. Similarly to the Mercedes dataset, the rental rates dataset contains relatively few features relative to its sample size and thus the multiscale method performs well. We note also that this in the first dataset that is both difficult to overfit and multidimensional. In this scenario, the mixture density network substantially underfits, likely due to the difficulty of accurately estimating the covariance matrices in a mixture of multivariate normals.

\section{Conclusion}
\label{sec:conclusion}
We have presented two approaches conditional density estimation that are fully nonparametric, scalable, make no independence assumptions, and leverage the underlying topological structure of the variables.  Our first approach, Multiscale Nets, effectively morphs density estimation into hierarchical classication, and it is designed to be a maximally flexible model for situations with a favorable ratio of the number of samples to the feature count. Our second approach, CDE trend filtering (TF), couples a multinomial model (like option 1 above) with a trend filtering penalty to smooth the underlying density estimate via an additional regularization term. This second approach works best in situations where data is sparser as a function of the number of features.  In each case, the features are mapped to raw logits---binomial in the multiscale case, multinomial in the CDE TF case---via an appropriate neural network. We presented extensive evidence that each of these two methods is superior to Gaussian mixture models (the current state of the art) in its appropriate domain.

\clearpage
\begin{small}
\bibliographystyle{abbrvnat}
\bibliography{arxiv}

\begin{thebibliography}{18}
\providecommand{\natexlab}[1]{#1}
\providecommand{\url}[1]{\texttt{#1}}
\expandafter\ifx\csname urlstyle\endcsname\relax
  \providecommand{\doi}[1]{doi: #1}\else
  \providecommand{\doi}{doi: \begingroup \urlstyle{rm}\Url}\fi

\bibitem[Bashtannyk and Hyndman(2001)]{bash:hynd:2001}
D.~M. Bashtannyk and R.~J. Hyndman.
\newblock Bandwidth selection for kernel conditional density estimation.
\newblock \emph{Computational Statistics and Data Analysis}, 36:\penalty0
  279--98, 2001.

\bibitem[Bishop(1994)]{bishop:mdn:1994}
C.~M. Bishop.
\newblock Mixture density networks.
\newblock 1994.

\bibitem[Carreau and Bengio(2009)]{carreau2009hybrid}
J.~Carreau and Y.~Bengio.
\newblock A hybrid pareto mixture for conditional asymmetric fat-tailed
  distributions.
\newblock \emph{Neural Networks, IEEE Transactions on}, 20\penalty0
  (7):\penalty0 1087--1101, 2009.

\bibitem[Fryzlewicz and Nason(2004)]{fryzlewicz:nason:2004}
P.~Fryzlewicz and G.~Nason.
\newblock A wavelet-{F}isz algorithm for {P}oisson intensity estimation.
\newblock \emph{Journal of Computational and Graphical Statistics},
  13:\penalty0 621--38, 2004.

\bibitem[Gal and Ghahramani(2015)]{gal2015dropout}
Y.~Gal and Z.~Ghahramani.
\newblock Dropout as a bayesian approximation: Representing model uncertainty
  in deep learning.
\newblock \emph{arXiv preprint arXiv:1506.02142}, 2015.

\bibitem[Jansen(2006)]{jansen:2006}
M.~Jansen.
\newblock Multiscale {P}oisson data smoothing.
\newblock \emph{Journal of the Royal Statistical Society (Series B)},
  68\penalty0 (1):\penalty0 27--48, 2006.

\bibitem[Lopes et~al.(2011)Lopes, McCulloch, and Tsay]{lopes2011cholesky}
H.~F. Lopes, R.~McCulloch, and R.~Tsay.
\newblock Cholesky stochastic volatility.
\newblock Technical report, August 2 2011, discussion paper, 2011.

\bibitem[Ma(2014)]{ma:2014}
L.~Ma.
\newblock Markov adaptive {P}\'olya trees and multi-resolution adaptive
  shrinkage in nonparametric modeling.
\newblock arXiv:1401.7241 [stat.ME], 2014.

\bibitem[Mauldin et~al.(1992)Mauldin, Sudderth, and
  Williams]{mauldin:sudderth:williams:1992}
R.~Mauldin, W.~Sudderth, and S.~Williams.
\newblock Polya trees and random distributions.
\newblock \emph{Annals of Statistics}, 20:\penalty0 1203--21, 1992.

\bibitem[Morin and Bengio(2005)]{morin2005hierarchical}
F.~Morin and Y.~Bengio.
\newblock Hierarchical probabilistic neural network language model.
\newblock In \emph{Aistats}, volume~5, pages 246--252. Citeseer, 2005.

\bibitem[Pu et~al.(2015)Pu, Yuan, Stevens, Li, and Carin]{pu2015deep}
Y.~Pu, X.~Yuan, A.~Stevens, C.~Li, and L.~Carin.
\newblock A deep generative deconvolutional image model.
\newblock \emph{arXiv preprint arXiv:1512.07344}, 2015.

\bibitem[Stone et~al.(2003)Stone, Schapire, Littman, Csirik, and
  McAllester]{stone:etal:2003}
P.~Stone, R.~E. Schapire, M.~L. Littman, J.~A. Csirik, and D.~McAllester.
\newblock Decision-theoretic bidding based on learned density models in
  simultaneous, interacting auctions.
\newblock \emph{Journal of Artificial Intelligence Research}, 19:\penalty0
  209--242, 2003.

\bibitem[Sugiyama et~al.(2010)Sugiyama, Takeuchi, Suzuki, Kanamori, Hachiya,
  and Okanohara]{sugiyama2010conditional}
M.~Sugiyama, I.~Takeuchi, T.~Suzuki, T.~Kanamori, H.~Hachiya, and D.~Okanohara.
\newblock Conditional density estimation via least-squares density ratio
  estimation.
\newblock In \emph{International Conference on Artificial Intelligence and
  Statistics}, pages 781--788, 2010.

\bibitem[Tsanas et~al.(2010)Tsanas, Little, McSharry, and
  Ramig]{dataset:parkinsons}
A.~Tsanas, M.~A. Little, P.~E. McSharry, and L.~O. Ramig.
\newblock Accurate telemonitoring of parkinson's disease progression by
  noninvasive speech tests.
\newblock \emph{Biomedical Engineering, IEEE Transactions on}, 57\penalty0
  (4):\penalty0 884--893, 2010.

\bibitem[Wang et~al.(2014)Wang, Sharpnack, Smola, and
  Tibshirani]{wang2014trend}
Y.-X. Wang, J.~Sharpnack, A.~Smola, and R.~J. Tibshirani.
\newblock Trend filtering on graphs.
\newblock \emph{arXiv preprint arXiv:1410.7690}, 2014.

\bibitem[Willett and Nowak(2007)]{willett:nowak:2007}
R.~Willett and R.~Nowak.
\newblock Multiscale {P}oisson intensity and density estimation.
\newblock \emph{{IEEE} Transactions on Information Theory}, 53\penalty0
  (9):\penalty0 3171--87, 2007.

\bibitem[Zen and Senior(2014)]{zen2014deep}
H.~Zen and A.~Senior.
\newblock Deep mixture density networks for acoustic modeling in statistical
  parametric speech synthesis.
\newblock In \emph{Acoustics, Speech and Signal Processing (ICASSP), 2014 IEEE
  International Conference on}, pages 3844--3848. IEEE, 2014.

\bibitem[Zhang et~al.(2016)Zhang, Isola, and Efros]{zhang2016colorful}
R.~Zhang, P.~Isola, and A.~A. Efros.
\newblock Colorful image colorization.
\newblock \emph{arXiv preprint arXiv:1603.08511}, 2016.

\end{thebibliography}
\end{small}

\end{document}